\documentclass[conference]{IEEEtran}
\usepackage{graphicx}

\usepackage{url}

\ifCLASSINFOpdf
\else
\fi
\begin{document}
\title{Incidental Scene Text Understanding:\\Recent Progresses on ICDAR 2015 Robust Reading Competition Challenge 4}

\author{\IEEEauthorblockN{Cong Yao, Jianan Wu, Xinyu Zhou, Chi Zhang, Shuchang Zhou, Zhimin Cao, Qi Yin}
\IEEEauthorblockA{Megvii Inc.\\
Beijing, 100190, China\\
Email: \{yaocong, wjn, zxy, zhangchi, zsc, czm, yq\}@megvii.com}}


%Telephone: (800) 555--1212, Fax: (888) 555--1212}


\maketitle

\begin{abstract}
%\boldmath
Different from focused texts present in natural images, which are captured with user's intention and intervention, incidental texts usually exhibit much more diversity, variability and complexity, thus posing significant difficulties and challenges for scene text detection and recognition algorithms. The ICDAR 2015 Robust Reading Competition Challenge 4 was launched to assess the performance of existing scene text detection and recognition methods on incidental texts as well as to stimulate novel ideas and solutions. This report is dedicated to briefly introduce our strategies for this challenging problem and compare them with prior arts in this field.
\end{abstract}


% \fi
\IEEEpeerreviewmaketitle

%%%%%%%%% BODY TEXT
%-------------------------------------------------------------------------
\section{Introduction}

In the past few years, scene text detection and recognition have drawn much interest and concern from both the computer vision community and document analysis community, and numerous inspiring ideas and effective approaches have been proposed~\cite{Ref:Chen2004, Ref:Epshtein2010, Ref:Neumann2010, Ref:Wang2011, Ref:Yao2012, Ref:Neumann2012, Ref:Bissacco2013, Ref:Yao2014, Ref:Yao2014C, Ref:Jaderberg2015B} to tackle these problems.

Though considerable progresses have been made by the aforementioned methods, it is still not clear that how these algorithms perform on incidental texts instead of focused texts. Incidental texts mean that texts appeared in natural images are captured without user's prior preference or intention and thus bear much more complexities and difficulties, such as blur, usual layout, non-uniform illumination, low resolution in addition to cluttered background.

The organizers of the ICDAR 2015 Robust Reading Competition Challenge 4~\cite{Ref:Karatzas2015} therefore prepared this contest to evaluate the performance of existing algorithms that were originally designed for focused texts as well as to stimulate new insights and ideas.

To tackle this challenging problem, we propose in this paper ideas and solutions that are both novel and effective. The experiments and comparisons on the ICDAR 2015 dataset evidently verify the effectiveness of the proposed strategies.

%-------------------------------------------------------------------------
\section{Dataset and Competition}  \label{Sec:Dataset}

The ICDAR 2015 dataset\footnote{\url{http://rrc.cvc.uab.es/?ch=4&com=downloads}} is from the Challenge 4 (Incidental Scene Text challenge) of the ICDAR 2015 Robust Reading Competition~\cite{Ref:Karatzas2015}. The dataset includes 1500 natural images in total, which are acquired using Google Glass. Different from the images from the previous ICDAR competitions~\cite{Ref:Lucas2005, Ref:Shahab2011, Ref:Karatzas2013}, in which the texts are well positioned and focused, the images from ICDAR 2015 are taken in an arbitrary or insouciance way, so the texts are usually skewed or blurred.

There are three tasks, namely Text Localization (Task 4.1), Word Recognition (Task 4.3) and End-to-End Recognition (Task 4.4), based on this benchmark. For details of the tasks, evaluations protocols and accuracies of the participating methods, refer to~\cite{Ref:Karatzas2015}.

%-------------------------------------------------------------------------
\section{Proposed Strategies}  \label{Sec:Strategies}

In this section, we will briefly describe the main ideas and work flows of the proposed strategies for text detection, word recognition and end-to-end recognition, respectively.

\subsection{Text Detection}  \label{Sec:Detection}

Most of the existing text detection systems~\cite{Ref:Chen2004, Ref:Wang2010, Ref:Epshtein2010, Ref:Neumann2011, Ref:Neumann2013B, Ref:Yao2014C, Ref:Jaderberg2014} detect text within local regions, typically through extracting character, word or line level candidates followed by candidate aggregation and false positive elimination, which potentially ignore the effect of wide-scope and long-range contextual cues in the scene. In this work, we explore an alternative approach and propose to localize text in a holistic manner, by casting scene text detection as a semantic segmentation problem.

\begin{figure}[!tp]
\centering
\includegraphics[width=0.95\linewidth]{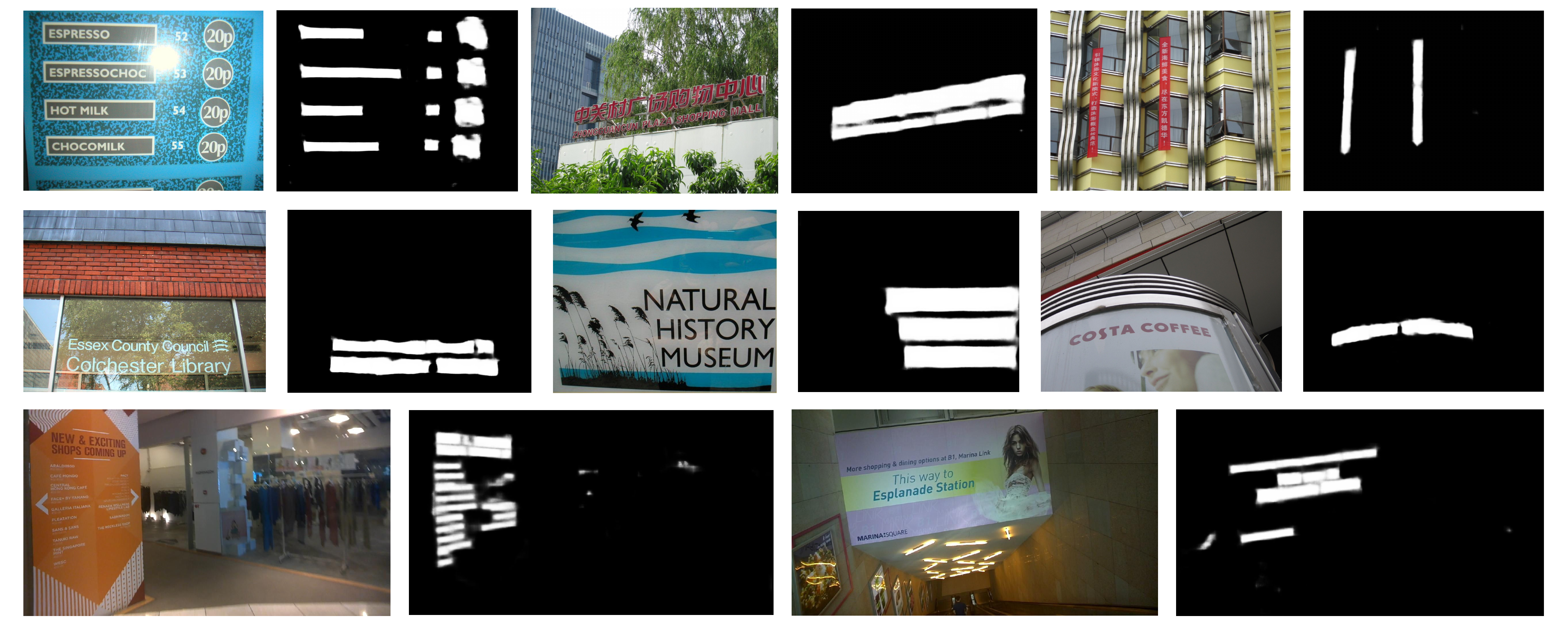}
\caption{Text regions predicted by the proposed text detection algorithm.}   \label{Fig:Headline}
\end{figure}

Specifically, we train a Fully Convolutional Networks (FCN)~\cite{Ref:Long2015} to perform per-pixel prediction on the probability of text regions (Fig.~\ref{Fig:Headline}). Detections are formed by subsequent thresholding and partition operations in the prediction map.

\subsection{Word Recognition}  \label{Sec:Recognition}

\begin{figure}[!tp]
\centering
\includegraphics[width=0.85\linewidth]{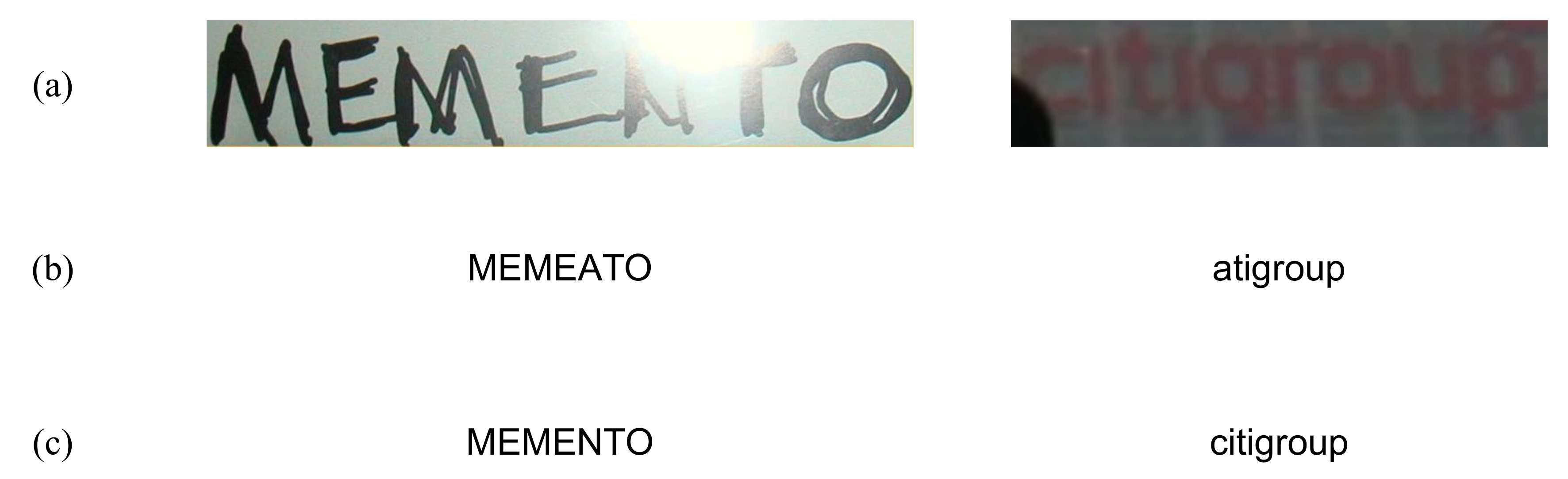}
\caption{Word recognition examples. (a) Original image. (b) Initial recognition result. (c) Recognition result after error correction.}   \label{Fig:Recognition}
\end{figure}

Word recognition is accomplished by training a combined model containing convolutional layers, and Long Short-Term Memory (LSTM)~\cite{Ref:Hochreiter1997} based Recurrent Neural Network (RNN) layers and a Connectionist Temporal Classification (CTC)~\cite{Ref:Graves2006} layer, followed by a dictionary based error correction (Fig.~\ref{Fig:Recognition}).

\subsection{End-to-End Recognition}  \label{Sec:End2End}

The method for end-to-end recognition is simply a combination the above two strategies. This combination has proven to be promising (see Sec.~\ref{Sec:Experiments} foe details).

%-------------------------------------------------------------------------
\section{Experiments and Comparisons}  \label{Sec:Experiments}

In this section, we will present the performances of the proposed strategies on the three tasks and compare them with the previous methods that have been evaluated on the ICDAR 2015 benchmark. All the results shown in this section can be also found on the homepage\footnote{\url{http://rrc.cvc.uab.es/?ch=4&com=evaluation}} of the ICDAR 2015 Robust Reading Competition Challenge 4.

\subsection{Text Localization (Task 4.1)}  \label{Sec:Task4.1}

\begin{table}
\caption{Detection Performances of different methods evaluated on ICDAR 2015.} 
\label{Tab:ICDAR2015Detection}
\begin{center}
\begin{tabular}{|c|c|c|c|}
\hline
\textbf{Algorithm}&\textbf{Precision}&\textbf{Recall}&\textbf{F-measure}  \\
\hline
\hline
Megvii-Image++                            &
0.724                                      &
\textbf{0.5696}                                      &
\textbf{0.6376}                                      \\
\hline
Stradvision-2~\cite{Ref:Karatzas2015}     &
\textbf{0.774}6                                      &
0.3674                                      &
0.4984                                      \\
\hline
Stradvision-1~\cite{Ref:Karatzas2015}     &
0.5339                                      &
0.4627                                      &
0.4957                                      \\
\hline
NJU~\cite{Ref:Karatzas2015}               &
0.7044                                      &
0.3625                                      &
0.4787                                      \\
\hline
AJOU~\cite{Ref:Koo2013}                   &
0.4726                                      &
0.4694                                      &
0.471                                      \\
\hline
HUST-MCLAB~\cite{Ref:Karatzas2015}        &
0.44                                      &
0.3779                                      &
0.4066                                      \\
\hline
Deep2Text-MO~\cite{Ref:Yin2015}           &
0.4959                                     &
0.3211                                      &
0.3898                                     \\
\hline
CNN MSER~\cite{Ref:Karatzas2015}          &
0.3471                                      &
0.3442                                      &
0.3457                                      \\
\hline
TextCatcher-2~\cite{Ref:Karatzas2015}     &
0.2491                                      &
0.3481                                     &
0.2904                                      \\
\hline
\end{tabular}
\end{center}
\vspace{-2mm}
\end{table}

The text detection performance of the proposed method (denoted as \textbf{Megvii-Image++}) as well as other competing methods on the Text Localization task are shown in Tab.~\ref{Tab:ICDAR2015Detection}. The proposed method achieves the highest recall (0.5696) and the second highest precision (0.724). Specifically, the F-measure of the proposed algorithm is significantly better than that of previous state-of-the-art (0.6376 vs. 0.4984). This confirms the effectiveness and advantage of the proposed approach.

Regarding running time, it takes the proposed text detection method about 20s to process a 640x480 image on CPU and 1s on GPU (no parallelization or multithread).

\subsection{Word Recognition (Task 4.3)}  \label{Sec:Task4.3}

\begin{table}
\caption{Word Recognition Performances of different methods evaluated on ICDAR 2015.} 
\label{Tab:ICDAR2015Word}
\begin{center}
\begin{tabular}{|c|c|c|c|c|}
\hline
\textbf{Algorithm}&\textbf{T.E.D.}&\textbf{C.R.W.}&\textbf{T.E.D.(upper)}&\textbf{C.R.W.(upper)}\\
\hline
\hline
Megvii-Image++                               &
\textbf{509.1}                                        &
\textbf{0.5782}                                       &
\textbf{377.9 }                                       &
\textbf{0.6399}                                        \\
\hline
MAPS~\cite{Ref:Kumar2012}                   &
1128.0                                        &
0.3293                                       &
1068.8                                      &
0.339                                        \\
\hline
NESP~\cite{Ref:Kumar2013}                   &
1164.6                                      &
0.3168                                       &
1094.9                                      &
0.3298                                      \\
\hline
DSM~\cite{Ref:Karatzas2015}                 &
1178.8                                      &
0.2585                                       &
1109.1                                      &
0.2797                                       \\
\hline
\end{tabular}
\end{center}
\vspace{-2mm}
\end{table}

Tab.~\ref{Tab:ICDAR2015Word} depicts the word recognition accuracies of our method and other participants on the Word Recognition task. As can be seen, our method substantially advances the state-of-the-art performance by nearly halving the Total Edit Distance (T.E.D.) and doubling the ratio of Correctly Recognized Words (C.R.W.). For the case insensitive settings, the  superiority of the proposed method over other competitors is also obvious.

\begin{table}
\caption{Word Recognition Performances of different methods evaluated on ICDAR 2013.} 
\label{Tab:ICDAR2013Word}
\begin{center}
\begin{tabular}{|c|c|c|c|c|}
\hline
\textbf{Algorithm}&\textbf{T.E.D.}&\textbf{C.R.W.}&\textbf{T.E.D.(upper)}&\textbf{C.R.W.(upper)}\\
\hline
\hline
Megvii-Image++                               &
\textbf{115.9}                                        &
\textbf{0.8283}                                       &
\textbf{94.1}                                       &
\textbf{0.8603}                                        \\
\hline
PhotoOCR~\cite{Ref:Bissacco2013}                   &
122.7                                        &
\textbf{0.8283}                                       &
109.9                                      &
0.853                                        \\
\hline
PicRead~\cite{Ref:Novikova2012}                   &
332.4                                        &
0.5799                                       &
290.8                                      &
0.6192                                        \\
\hline
NESP~\cite{Ref:Kumar2013}                   &
360.1                                      &
0.642                                       &
345.2                                      &
0.6484                                      \\
\hline
PLT~\cite{Ref:Karatzas2013}                   &
392.1                                      &
0.6237                                       &
375.3                                      &
0.6311                                      \\
\hline
MAPS~\cite{Ref:Kumar2012}                   &
421.8                                        &
0.6274                                        &
406                                      &
0.6329                                        \\
\hline
PIONEER~\cite{Ref:Weinman2013}                   &
479.8                                        &
0.537                                        &
426.8                                      &
0.5571                                        \\
\hline
\end{tabular}
\end{center}
\vspace{-2mm}
\end{table}

To further verify the effectiveness of the proposed strategy for word recognition, we also evaluated it on the test set o from the Word Recognition task of ICDAR 2013. As can be seen from Tab.~\ref{Tab:ICDAR2013Word}, the proposed method for word recognition outperforms the previous state-of-the-art algorithm PhotoOCR~\cite{Ref:Bissacco2013} as well as other competitors, in all metrics.

\subsection{End-to-End Recognition (Task 4.4)}  \label{Sec:Task4.4}

\begin{table*}
\caption{End-to-End Recognition Performances of Different Methods Evaluated on ICDAR 2015 Challenge 4.} 
\label{Tab:ICDAR2015End2End}
\begin{center}
\begin{tabular}{|c|ccc|ccc|ccc|}
\hline
\textbf{Algorithm}&&\textbf{Strong}&&&\textbf{Weak}&&&\textbf{Generic}&\\
&\textbf{P}&\textbf{R}&\textbf{F}&\textbf{P}&\textbf{R}&\textbf{F}&\textbf{P}&\textbf{R}&\textbf{F}\\
\hline
\hline
Megvii-Image++                               &
0.5748                                        &
0.3938                                        &
\textbf{0.4674}                                       &
\textbf{0.4919}                                        &
\textbf{0.337}                                        &
\textbf{0.4}                                       &
\textbf{0.4041}                                         &
\textbf{0.2768}                                         &
\textbf{0.3286}                                        \\
\hline
Stradvision-2~\cite{Ref:Karatzas2015}                              &
\textbf{0.6792}                                       &
0.3221                                         &
0.4370                                         &
-                                        &
-                                       &
-                                        &
-                                        &
-                                       &
-                                       \\
\hline
Baseline-TextSpotter~\cite{Ref:Neumann2013}                               &
0.6221                                       &
0.2441                                        &
0.3506                                        &
0.2496                                        &
0.1656                                        &
0.1991                                        &
0.1832                                        &
0.1358                                        &
0.1560                                        \\
\hline
StradVision\_v1~\cite{Ref:Karatzas2015}                               &
0.2851                                        &
\textbf{0.3977}                                         &
0.3321                                        &
-                                        &
-                                       &
-                                        &
-                                        &
-                                       &
-                                       \\
\hline
NJU Text (Version3)~\cite{Ref:Karatzas2015}                               &
0.488                                        &
0.2451                                         &
0.3263                                        &
-                                        &
-                                       &
-                                        &
-                                        &
-                                       &
-                                       \\
\hline
Beam search CUNI~\cite{Ref:Karatzas2015}                               &
0.3783                                       &
0.1565                                        &
0.2214                                         &
0.3372                                       &
0.1401                                         &
0.1980                                        &
0.2964                                       &
0.1237                                        &
0.1746                                        \\
\hline
Deep2Text-MO~\cite{Ref:Yin2015}                               &
0.2134                                        &
0.1382                                        &
0.1677                                        &
0.2134                                        &
0.1382                                        &
0.1677                                        &
0.2134                                        &
0.1382                                        &
0.1677                                         \\
\hline
Baseline (OpenCV+Tesseract)~\cite{Ref:Gomez2014}                               &
0.409                                       &
0.0833                                        &
0.1384                                       &
0.3248                                       &
0.0737                                        &
0.1201                                       &
0.1930                                       &
0.0506                                        &
0.0801                                        \\
\hline
Beam search CUNI+S~\cite{Ref:Karatzas2015}                               &
0.8108                                       &
0.0722                                        &
0.1326                                        &
0.0592                                        &
0.6474                                        &
0.1085                                        &
0.0380                                         &
0.3496                                       &
0.0686                                        \\
\hline
\end{tabular}
\end{center}
\vspace{-2mm}
\end{table*}

The end-to-end recognition performances of different methods on the End-to-End Recognition task are demonstrated in Tab.~\ref{Tab:ICDAR2015End2End}. For the Strongly Contextualised setting, the proposed method achieves the best F-measure (0.4674) and the second best in recall (0.3938). For the Weakly Contextualised and Generic settings, which are more close to real-world applications and more realistic, the proposed strategy obtains overwhelmingly superior accuracies than the existing methods, almost doubling all the metrics (precision=0.4919, recall= 0.337, F-measure=0.4 for the Weakly Contextualised setting and precision=0.4041, recall= 0.2768, F-measure=0.3286 for the Generic setting).

\begin{table*}
\caption{End-to-End Recognition Performances of Different Methods Evaluated on ICDAR 2015 Challenge 1 (Born-Digital).} 
\label{Tab:ICDAR2015End2EndCH1}
\begin{center}
\begin{tabular}{|c|ccc|ccc|ccc|}
\hline
\textbf{Algorithm}&&\textbf{Strong}&&&\textbf{Weak}&&&\textbf{Generic}&\\
&\textbf{P}&\textbf{R}&\textbf{F}&\textbf{P}&\textbf{R}&\textbf{F}&\textbf{P}&\textbf{R}&\textbf{F}\\
\hline
\hline
Megvii-Image++                               &
\textbf{0.9253}                                     &
\textbf{0.7921}                                        &
\textbf{0.8535}                                       &
\textbf{0.9059}                                    &
\textbf{0.7900}                                        &
\textbf{0.8440}                                       &
0.8331                                         &
\textbf{0.7497}                                         &
\textbf{0.7892}                                        \\
\hline
Deep2Text II+~\cite{Ref:Yin2015}                              &
0.9227                                        &
0.7392                                         &
0.8208                                         &
0.8916 	                                         &
0.7378                                       &
0.8075                                        &
\textbf{0.8532}	                                         &
0.7316                                       &
0.7877                                       \\
\hline
Stradvision-2~\cite{Ref:Karatzas2015}                              &
0.8393 	                                      &
0.7302                                         &
0.7810                                         &
0.7761                                     &
0.7086                                       &
0.7408                                        &
0.5735                                       &
0.5668                                       &
0.5701                                       \\
\hline
Deep2Text II-1~\cite{Ref:Yin2015}                              &
0.8097                                       &
0.7337                                         &
0.7698                                        &
0.8097                                     &
0.7337                                       &
0.7698                                        &
0.8097	                                         &
0.7337                                       &
0.7698                                       \\
\hline
StradVision-1~\cite{Ref:Karatzas2015}                               &
0.8472                                       &
0.7017                                        &
0.7676                                        &
0.7890                                        &
0.6787                                       &
0.7297                                        &
0.5820                                       &
0.5431                                       &
0.5619                                       \\
\hline
Deep2Text I~\cite{Ref:Yin2015}                              &
0.8346	                                      &
0.6140                                         &
0.7075                                        &
0.8346                                     &
0.6140                                       &
0.7075                                        &
0.8346                                       &
0.6140                                       &
0.7075                                       \\
\hline
PAL (v1.5)~\cite{Ref:Karatzas2015}                              &
0.6522                                     &
0.6154                                         &
0.6333                                        &
-                                     &
-                                       &
-                                        &
-                                       &
-                                       &
-                                       \\
\hline
NJU Text (Version3)~\cite{Ref:Karatzas2015}                              &
0.6012                                     &
0.4131                                         &
0.4897                                        &
-                                     &
-                                       &
-                                        &
-                                       &
-                                       &
-                                       \\
\hline
Baseline OpenCV 3.0 + Tesseract~\cite{Ref:Karatzas2015}                              &
0.4648	                                     &
0.3713                                         &
0.4128                                        &
0.4720                                  &
0.3282                                       &
0.3872                                        &
0.3029                                    &
0.2420                                     &
0.2690                                      \\
\hline
\end{tabular}
\end{center}
\vspace{-2mm}
\end{table*}

We have also assessed the proposed system on the dataset of the ICDAR 2015 Robust Reading Competition Challenge 1 (Born-Digital). The end-to-end recognition performances of different algorithms on the End-to-End Recognition task are demonstrated in Tab.~\ref{Tab:ICDAR2015End2EndCH1}. As can be observed, on the dataset of Challenge 1, where all the text are born-digital, the proposed method achieves state-of-the-art performance as well.

Overall, the significantly improved performances on the three tasks evidently prove the effectiveness and superiority of the proposed strategies.

%-------------------------------------------------------------------------
\section{Conclusions}  \label{Sec:Conclusions}

In this paper, we have presented our strategies for incidental text detection and recognition in natural scene images. The strategies introduce novel insights on the problem and exploit the power of deep learning~\cite{Ref:Krizhevsky2012}. The experiments on the benchmark of the ICDAR 2015 Robust Reading Competition Challenge 4 as well as Challenge 1 demonstrate that the proposed strategies lead to substantially enhanced performance than previous state-of-the-art approaches.


%\section*{Acknowledgment}




\bibliographystyle{IEEEtran}
\bibliography{benchmark}


%\begin{thebibliography}{1}


%\end{thebibliography}

\end{document}